\documentclass{article} 
\usepackage[preprint]{neurips_2025}

\usepackage[utf8]{inputenc} 
\usepackage[T1]{fontenc}    
\usepackage{hyperref}       
\usepackage{url}            
\usepackage{booktabs}       
\usepackage{amsfonts}       
\usepackage{nicefrac}       
\usepackage{microtype}      
\usepackage{xcolor}         

\usepackage{amsmath,amsfonts,bm}

\def\eqref#1{equation~\ref{#1}}

\def\1{\bm{1}}

\DeclareMathAlphabet{\mathsfit}{\encodingdefault}{\sfdefault}{m}{sl}
\SetMathAlphabet{\mathsfit}{bold}{\encodingdefault}{\sfdefault}{bx}{n}

\usepackage{graphicx}
\usepackage{wrapfig}
\usepackage{subfigure}
\usepackage{float}
\usepackage{booktabs} 
\usepackage{multirow} 
\usepackage{makecell} 
\usepackage{caption} 

\usepackage{hyperref}
\usepackage{url}
\usepackage{cleveref}

\title{A Pretrained Probabilistic Transformer for City-Scale Traffic Volume Prediction}

\author{%
  Shiyu Shen \\
  Nankai University
  \And
  Bin Pan \\
  Nankai University
  \And
  Guirong Xue \\
  Zhejiang lab
}

\begin{document}

\maketitle

\begin{abstract}
    City-scale traffic volume prediction plays a pivotal role in intelligent transportation systems, yet remains a challenge due to the inherent incompleteness and bias in observational data. Although deep learning-based methods have shown considerable promise, most existing approaches produce deterministic point estimates, thereby neglecting the uncertainty arising from unobserved traffic flows. Furthermore, current models are typically trained in a city-specific manner, which hinders their generalizability and limits scalability across diverse urban contexts. To overcome these limitations, we introduce TrafficPPT, a Pretrained Probabilistic Transformer designed to model traffic volume as a distributional aggregation of trajectories. Our framework fuses heterogeneous data sources—including real-time observations, historical trajectory data, and road network topology—enabling robust and uncertainty-aware traffic inference. TrafficPPT is initially pretrained on large-scale simulated data spanning multiple urban scenarios, and later fine-tuned on target cities to ensure effective domain adaptation. Experiments on real-world datasets show that TrafficPPT consistently surpasses state-of-the-art baselines, particularly under conditions of extreme data sparsity. Code will be open.
\end{abstract}

\section{Introduction}

Traffic volume prediction serves as a fundamental task in urban traffic management, providing essential insights for congestion control, safety improvement, and infrastructure development. Existing approaches encompass time series models \cite{vlahogianni2014short}, deep learning architectures \cite{lv2014traffic}, and graph neural networks \cite{yu2017spatio,li2017diffusion}.

Despite recent advances, many existing methods assume access to complete traffic datasets \cite{fang2020meta, chen2024traffic}, which significantly limits their applicability in real-world scenarios. In practice, traffic monitoring relies primarily on two complementary yet imperfect data sources: (1) GPS data, which provides relatively detailed trajectory-level information but only for a subset of vehicles, and (2) fixed road sensors, which capture all passing vehicles but are sparsely deployed, often limited to key intersections or strategic checkpoints. This results in spatially fragmented observations, where urban centers typically exhibit high sensor density while suburban and rural areas remain under-monitored. Even in advanced cities with extensive downtown coverage, the intrinsic sparsity and uneven distribution of monitoring infrastructure still pose fundamental challenges.

To overcome these limitations, some researchers have shifted toward utilizing incomplete checkpoint data as a more practical solution for city-scale prediction \cite{chen2023vehicle}. Current approaches can be broadly categorized into: (1) traditional methods that employ prior probabilities for traffic volume estimation \cite{yu2023city,bao2023pket}, and (2) deep learning techniques that reconstruct missing trajectories from partial observations \cite{zhang2019short,guo2024m}. While demonstrating promising results, these methods exhibit two shortcomings: First, their end-to-end architectures lack interpretability while failing to account for potential alternative scenarios. Second, the mainstream city-specific training paradigm limits model generalizability and practical deployment potential. This represents a fundamental constraint, as most cities lack sufficient training data despite sharing common traffic pattern characteristics across urban areas.

To address these challenges, we propose a pretrained probabilistic framework that provides a comprehensive and universal solution for traffic volume prediction. Our approach fundamentally reformulates the problem by treating traffic volume as an aggregation of probabilistic trajectory distributions. Rather than determining a deterministic trajectory for each vehicle, we estimate the probability of its presence on any road at any time. These probabilistic distributions are estimated through a neural network, which follows a two-stage training paradigm: (1) a pretraining stage on simulated data on diverse cities to learn universal traffic patterns, followed by (2) a fine-tuning stage on city-specific data to adapt to local road network characteristics.

Building upon the probabilistic pretraining framework, we develop TrafficPPT, a transformer-based architecture that achieves comprehensive traffic volume inference with three innovations. First, the model integrates multimodal urban data - including real-time observations, historical trajectories, and road network topology - by projecting all heterogeneous inputs into a unified latent representation space. Second, we design a multi-view attention mechanism that dynamically captures spatial-temporal dependencies to estimate vehicle trajectory probabilities across the road network. Third, the architecture enables direct parallel prediction of complete traffic volumes through non-autoregressive inference, satisfying the strict latency requirements of real-world deployment scenarios. Experiments on city-scale road networks demonstrates that TrafficPPT outperforms existing methods in both accuracy and efficiency. Specifically, the model exhibits particular robustness to data sparsity, maintaining accurate predictions even with highly incomplete observations. The primary contributions of this paper are as follows:

\begin{itemize}
    \item \textbf{Probabilistic Transformer Architecture:} We propose TrafficPPT, which integrates multi-source traffic data to probabilistically infer city-scale traffic volumes. This framework captures the uncertainty of missing data and provides a more comprehensive inference.
    \item \textbf{Generalizable Pretraining Framework:} We introduce a two-stage training paradigm combining large-scale simulation pretraining with targeted fine-tuning. The pretraining learns universal traffic patterns from diverse synthetic environments, enabling adaption to specific cities with limited data.
    \item \textbf{Data-Efficient Performance:} TrafficPPT demonstrates high efficiency and high tolerance to missing data. With only 20\% observations, TrafficPPT outperforms other models that require 50\% observation ratio. These properties enable practical applications in resource-constrained urban settings.
\end{itemize}

\section{Related Work}

\subsection{Traffic Volume Prediction}

Traffic volume prediction has traditionally relied on historical trajectory data collected through infrastructure-based sensors or GPS services, using models like LSTMs and GNNs \cite{yu2017spatio,li2017diffusion,zhang2017deep,diao2019dynamic}. However, the requirement for dense and citywide data collection renders these methods impractical in many real-world scenarios. To address this challenge, some studies focus on checkpoint-based data from key intersections, offering a more accessible alternative \cite{liu2020dynamic,kalander2020spatio,liu2018urban}. A common line of work relies on prior distributions, assuming that missing trajectories follow specific patterns such as the shortest route \cite{patterson2020learning,hunter2013path,iio2023distribution}. However, this approach often fail to reflect real-world driving behaviors. Other approaches adopt deep learning-based reconstruction. Although effective in certain settings, these models often neglect the stochastic and context-dependent vehicle behaviors. Moreover, both statistical and neural models frequently presume a single-path assumption, overlooking the multimodal nature of vehicle routing decisions \cite{tang2023explainable}. As a result, their predictive capabilities degrade in highly sparse observation scenarios, constraining their applicability to city-scale deployment. Finally, while time-series models have demonstrated strong performance in capturing temporal dependencies, they often suffer from cumulative prediction errors and high inference latency, posing scalability challenges in large urban environments \cite{ma2015long,zhao2019t,wong2022view,huang2023reconstructing}.

\subsection{Pretrained Transformer}

Pretrained transformers have gained significant attention following the success in natural language processing \cite{achiam2023gpt,liu2024deepseek}. Two predominant pretraining paradigms have emerged: (1) Masked Language Modeling, which randomly masks a subset of input tokens and trains the model to reconstruct the masked content \cite{devlin2019bert}; (2) Next Token Prediction, which conditions on a sequence of preceding tokens to predict the subsequent one \cite{radford2018improving}. These strategies have been effectively applied to various tasks, including computer vision \cite{he2022masked} and graph representation learning \cite{hu2019strategies,hu2020gpt}. In the context of traffic prediction, some studies have explored the use of pretrained transformers \cite{hou2024masked,zhou2024trafficformer,jin2021trafficbert}. However, these methods often focus on specific datasets with sufficient data and lack a comprehensive framework that integrates heterogeneous information.

\section{Method}

In this section, we present our model TrafficPPT. We begin by introducing the foundational notations in \cref{problem_setup}. The probabilistic modeling framework is detailed in \cref{modeling}, including the formulation of the probabilistic objective, estimation procedures, and inference strategies. Then, we describe the model architecture of TrafficPPT, with additional implementation details provided in Appendix \ref{model_structure_}. Finally, we show the pretraining and fine-tuning mechanisms in \cref{pretraining_finetuning}.

\subsection{Problem Setup\label{problem_setup}}

\begin{figure*}[htbp]
    \centering
    \includegraphics[width=0.55\linewidth]{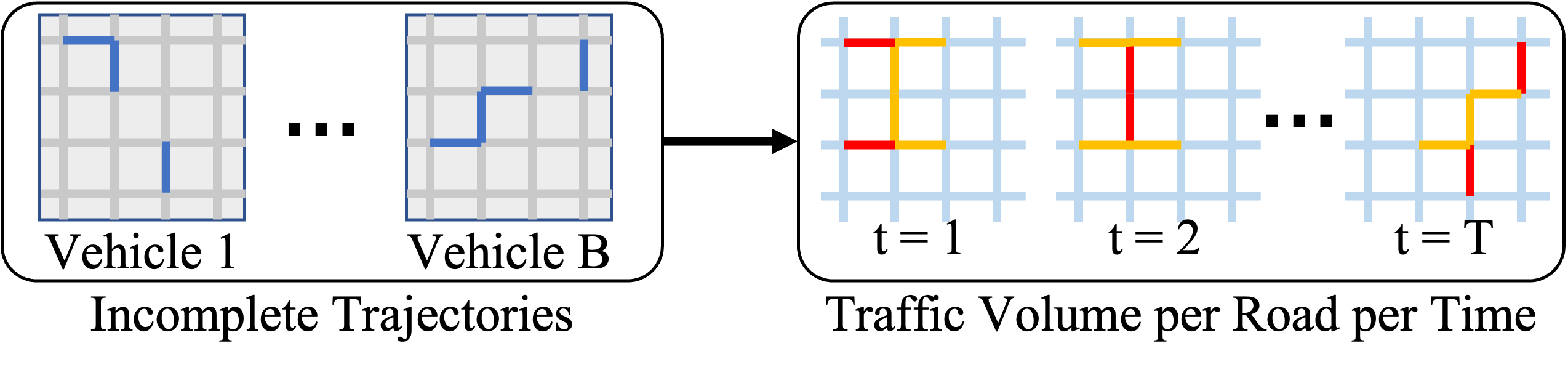}
    \caption{Objective of city-scale traffic volume prediction. The incomplete trajectories of each vehicle are collected from road sensors. We aim to predict vehicle counts per road segment at each time step throughout the city.\label{fig_setup}}
\end{figure*}

We represent the road network as a directed graph $\mathbb{G} = (\mathbb{V}, \mathbb{E})$, where $\mathbb{V} = \{1, 2, \dots, V\}$ denotes the set of nodes (i.e., intersections), and $\mathbb{E} = \{e_1, e_2, \dots, e_E\}$ denotes the set of edges (i.e., roads). Each edge $e_i = (o_i, d_i, l_i)$ is defined by its origin node $o_i \in \mathbb{V}$, destination node $d_i \in \mathbb{V}$, and weight $l_i$, which can represent various road attributes such as length, average speed, or traffic intensity.

Given that the number of nodes is typically much smaller than the number of edges, we represent vehicle trajectories as sequences of nodes. Specifically, a trajectory is denoted by $\mathbf{x} = [v_1, v_2, \dots, v_T]$, where $v_t \in \{0, 1, \dots, V\}$, and $T$ is the maximum number of time steps. The special token $v_t = 0$ indicates that the vehicle is not observable (e.g., due to occlusion or sensor limitations). To encode travel time along each edge, node entries may be repeated in the sequence. For example, the trajectory $[1, 2, 2, 2, 2, 3, 4, 0]$ indicates that the vehicle spends 1 step on edge $(1,2)$, 4 steps on edge $(2,3)$, 1 step on edge $(3,4)$, and subsequently disappears. In real-world scenarios, observed trajectories are often incomplete. For instance, if only nodes 1 and 3 are observable, the corresponding observation would be $[1, 0, 0, 0, 0, 3, 0, 0]$. 

Let $X \in \{0, 1, \dots, V\}^{B \times T}$ denote the set of $B$ observed trajectories over $T$ time steps. The goal of traffic volume prediction is to estimate the number of vehicles traversing each road segment at every time step. Formally, we define the traffic volume tensor as $\mathbf{Vol} \in \mathbb{R}^{E \times T}$, where $\mathbf{Vol}[i, t]$ denotes the number of vehicles passing through edge $e_i$ at time $t$.

\subsection{Probabilistic Modeling for Traffic Volume Prediction\label{modeling}}

We model the movement of each vehicle as an independent probabilistic process over the road network. Specifically, we define the trajectory probability tensor as $Y \in [0,1]^{B \times T \times V}$, where $Y[b, t, v]$ denotes the probability that the $b$-th vehicle occupies node $v$ at time step $t$. We use $q_\theta(Y|X,X_{his}, A)$ to approximate trajectory probability, where $\theta$ is the model parameters, $X_{his} \in \{0,1,...,V\}^{B,N,T}$ is the historical trajectory information. The road network is encoded via a set of adjacency tables $A = \{A_0, A_1, \dots, A_M\}$, where $M$ denotes the number of feature-specific adjacency matrices. In particular, $A_0 \in \{0, 1, \dots, V\}^{B \times V \times L}$ encodes the topological graph, with $L$ indicating the maximum connectivity. The remaining tables provide auxiliary features such as speed limits, road lengths, and road categories. Given access to the ground truth trajectory distribution $p(Y)$, the model is trained by minimizing KL-divergence between $p$ and $q_\theta$. To be specific:

\begin{align}
    \mathcal{L} = \sum_{b=1}^{B}\sum_{t=1}^{T}\sum_{v=1}^{V}p(Y[b,t,v])\log\frac{p(Y[b,t,v])}{q_\theta(Y[b,t,v]|X[b],X_{hist}[b],A[b])}
\end{align}
where $p(Y[b,t,v])$ and $q_\theta(Y[b,t,v]|X[b],X_{hist}[b],A[b])$ are both Bernoulli distributed.

We train TrafficPPT using complete trajectories. We mask the complete trajectories by different strategies in the pretraining and fine-tuning stages. When $N+1$ trajectories are available for a given vehicle, we randomly designate one as the observation and use the remaining as historical data. Since the ground truth trajectory is complete, the probability collapses into a one-hot distribution: $p(Y[b,t,v]) = 1$ if $v = X[b,t]$, otherwise $p(Y[b,t,v]) = 0$. Consequently, the loss function simplifies to the cross-entropy loss. The final loss function is calculated as follows:
\begin{align}
    \mathcal{L}([X,X_{hist},A],\tilde{Y}) = -\sum_{b=1}^{B}\sum_{t=1}^{T}\sum_{v=1}^{V}\tilde{Y}[b,t,v]\log q_\theta(Y[b,t,v]|X[b],X_{hist}[b],A[b])\label{loss_function}
\end{align}
where $\tilde{Y}[b,t,v]$ is one-hot encoding of the ground truth trajectory.

The traffic volume is estimated as the expectation over the predicted trajectory distributions. First, the node-level trajectory probability is converted into edge-level trajectory probability by the multiplication of the origin and destination node probabilities. To be specific:
\begin{align}
    \dot{Y}[b,t,i] = Y[b,t,o_i]\times Y[b,t+1,d_{i}] + Y[b,t,o_{i}]\times Y[b,t+1,o_{i}]
\end{align}
where $\dot{Y}[b,t,i]$ denotes the probability that vehicle $b$ occupies edge $e_i$ at time $t$. The formulation captures two scenarios: (1) the vehicle moves from node $o_i$ to $d_i$ (transition), and (2) the vehicle remains at node $o_i$ (dwell), both contributing to the edge occupancy. The traffic volume can be estimated by summing the expected contributions of all vehicles. To be specific:
\begin{align}
    \mathbf{Vol}[i,t] = \sum_{b=1}^{B}(\frac{\dot{Y}[b,t,i]}{\sum_{j=0}^{E}\dot{Y}[b,t,j]})
\end{align}
where $\mathbf{Vol}[i,t]$ is the volume of road $i$ at time $t$. We apply normalization to make sure that the total volume contribution from each car sums to 1.

\subsection{Model Architecture\label{overall_arch}}
\begin{figure*}[htbp]
    \centering
    \includegraphics[width=0.95\linewidth]{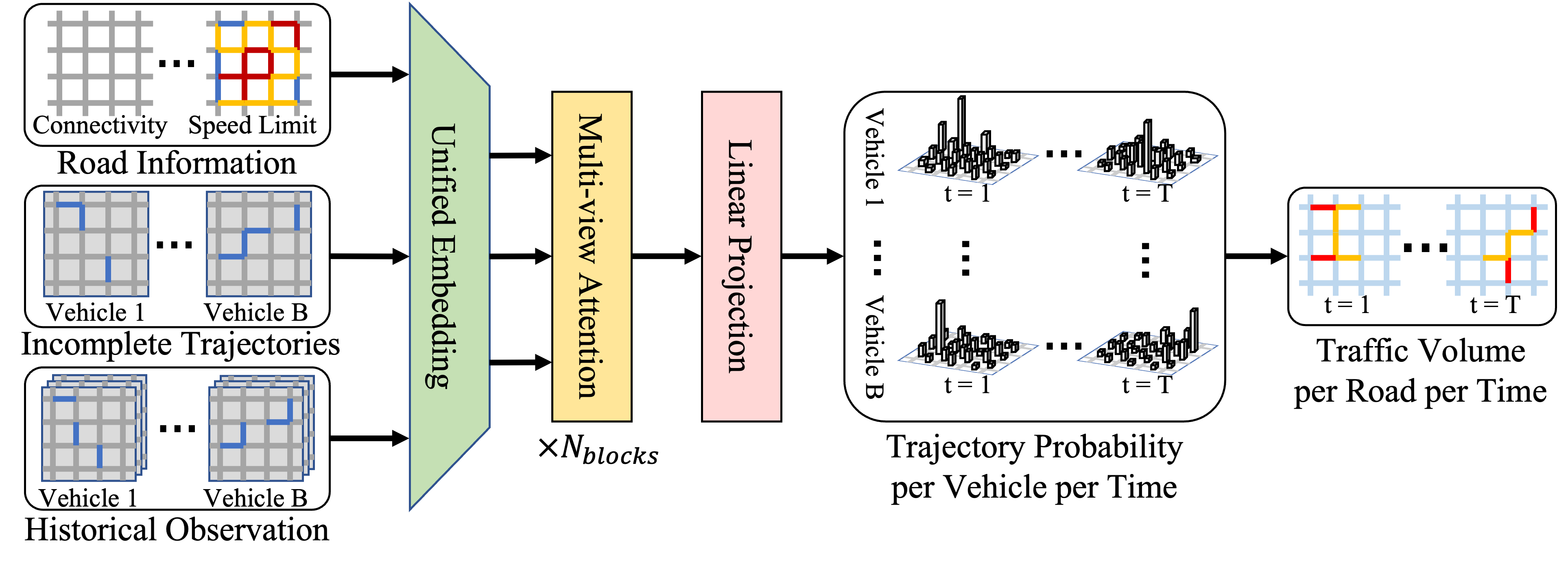}
    \caption{Overall architecture of the proposed TrafficPPT. \label{fig_overall_arch}}
\end{figure*}

The overall architecture of our model is illustrated in \cref{fig_overall_arch}. It comprises three components: embedding layers, multi-view attention blocks and a linear projection head. The embedding layers transform heterogeneous inputs—including observed trajectories, historical trajectories, and road network attributes—into a shared latent representation space. These embeddings are then fused via the multi-view attention blocks, which model complex cross-modal interactions. Finally, a linear projection layer followed by a softmax activation outputs the trajectory probability distributions, which are subsequently used to infer traffic volume. Further architectural details are provided in Appendix \ref{model_structure_}.

The embedding module contains three parallel branches: observation embedding, history embedding, and road network embedding. For a given observed trajectory $X$, we first map each discrete node index to a dense embedding via a learnable matrix $E_{\text{traj}}$. These embeddings are further refined by an MLP to produce the final observation tokens. Since historical data share the same structural format as observations, we reuse the same embedding pipeline. For the road network information, each adjacency is processed differently depending on its modality. Discrete attributes are embedded using a trainable embedding matrix $E_{\text{adj}}$, while continuous features are encoded through an MLP. To manage computational overhead, we apply average pooling over adjacency tokens to derive compact node-level representations.

The multi-view attention block integrates information from observations, historical data, and the road network via cross-attention mechanisms. The adjacency tokens represent the information of each node, which is comprehensive but not efficient. To address this problem, we adopt the multi-query attention mechanism \cite{ainslie2023gqa} to reduce the computation. As demonstrated in our ablation study, the multi-query attention effectively reduces computation with minimal impact on model performance. After the cross-attention, we apply a self-attention and a feed-forward network with residual link to enhance representational capacity. The multi-view attention module is stacked $N_{\text{block}}$ times to capture deep, hierarchical dependencies across modalities.

\subsection{Pretraining and Fine-tuning Mechanism.\label{pretraining_finetuning}}

\begin{figure*}[htbp]
    \centering
    \includegraphics[width=0.9\linewidth]{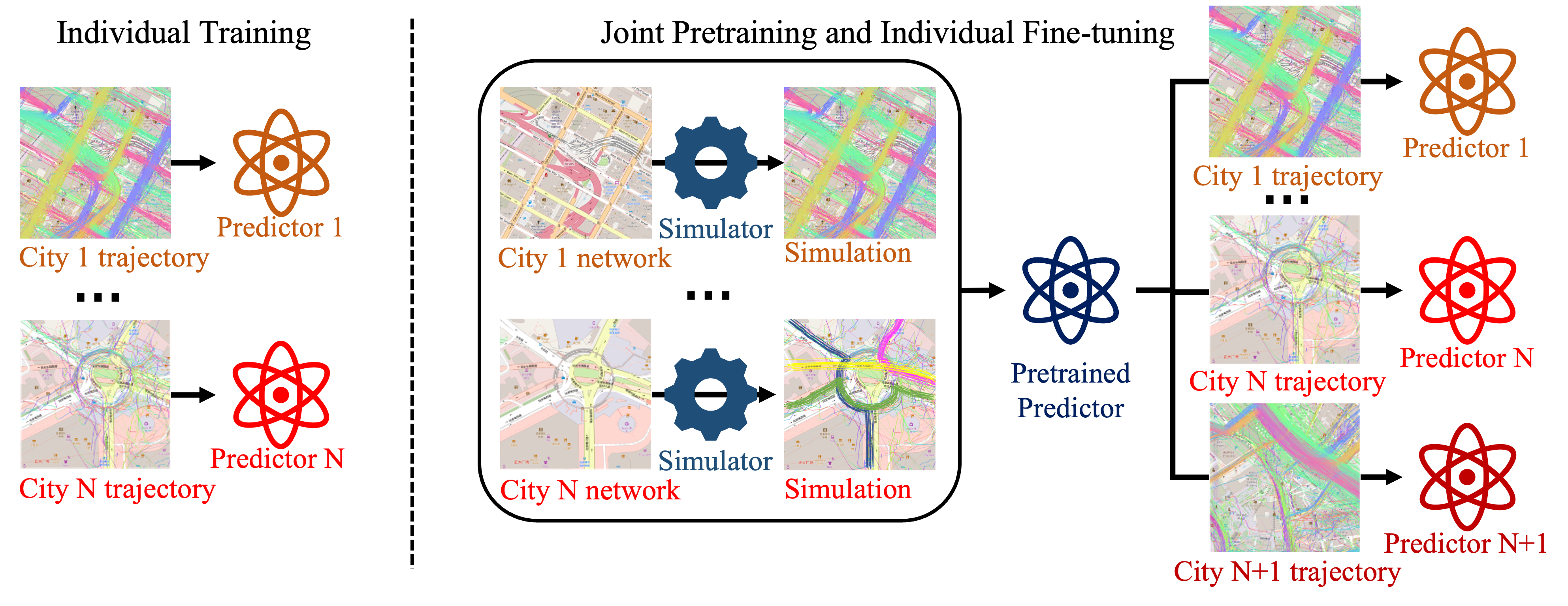}
    \caption{Pretraining and fine-tuning mechanism of TrafficPPT. The simulator can generate trajectories for cities with insufficient data, and the pretrained model can adapt to unseen cities.\label{fig_pretraining_finetuning}}
\end{figure*}
To improve the model's generalization ability, we adopt a pretraining and fine-tuning mechanism. The pretraining stage is conducted on simulation data, consisting of millions of complete trajectories from different cities. Road networks for these cities are collected from OpenStreetMap \cite{openstreetmap,boeing2017osmnx}, which provide information on connectivity, road types, segment lengths, speed limits, and other relevant attributes. For each city, we randomly assign origins and destinations for one million vehicles and use a simulator to generate the corresponding trajectories. The simulator can be either a vanilla version or an advanced version. The vanilla simulator generates trajectories based on the shortest paths, while allowing road weights to be adjusted according to length, speed limits, or other attributes to emulate various scenarios. The advanced simulator \cite{liang2023cblab} constructs virtual environments by placing all vehicles onto the road network and simulating the traffic flow dynamics. For each vehicle, we generate $N+1$ trajectories, corresponding to one observation trajectory and $N$ historical trajectories. These trajectories are then used to train TrafficPPT, following a strategy similar to BERT \cite{devlin2019bert}. Specifically, we randomly mask the trajectories with a ratio of $1-\alpha$ and train TrafficPPT to recover the complete trajectories.

The fine-tuning stage is performed on real-world data in a city-specific manner. Unlike the pretraining stage, we randomly select observable checkpoints from the city intersections with a ratio of $\alpha$, and consistently mask the unobserved checkpoints. Additionally, the final step of each observation is masked to enhance predictive performance. For each vehicle, if the dataset contains $N+1$ trajectories, we randomly select one as the observation; otherwise, we resample the available data to construct $N+1$ trajectories. Similarly, the training objective is to recover the masked portions of the trajectory.

\section{Experiments\label{experiments}}
In this section, we evaluate the proposed method on real-world road networks. We begin by describing the experiment setup in \cref{experiment_settings}, including the data description and model details. Next, we present the main results in \cref{main_results}, offering both an overall comparison and an in-depth performance analysis. Subsequently, we show the effectiveness of the pretraining and fine-tuning mechanism in \cref{pretraining_finetuning_analysis} Finally, we conduct an ablation study in \cref{ablation_study} to assess the impact of different computation-efficient mechanisms.

\subsection{Experiment Settings\label{experiment_settings}}
\subsubsection{Data Description\label{data_description}}
The experiment is conducted on two real-world cities: Boston and Jinan. The Boston road network comprises 241 nodes and 369 edges. We use simulator to generate trajectories. The maximum time step is set to 60. In total, 500,000 trajectories are simulated for training and 10,000 for testing. The weighted adjacency matrix used in each simulation is recorded as the road network input. The Jinan road network data is obtained from \cite{yu2023city}, consisting of 8,908 nodes and 23,312 edges. In addition to position and connectivity information, this dataset includes road length, road type, and complete trajectories of 963,125 individual vehicles. We randomly select 800,000 trajectories for training, with the remainder used for testing. We use road length as the road weight for the road network inputs. Since the temporal scale of trajectories varies widely—from seconds to hours—we standardize the data by rescaling all trajectories to 60 time steps. For each vehicle, we randomly select 1 trajectory as the observation and 4 trajectories as historical inputs.

\subsubsection{Training Details\label{training_details}}

For fair comparison, we use only the vanilla simulator during pretraining and do not include data from additional cities. The models for different cities share the same transformer backbone but have separate embedding layers. For the Boston we utilize MLP as the tokenizer for better performance. For the large Jinan road network we apply the discretization mechanism to reduce computational overhead. We simulate 1,000,000 samples for each dataset, ensuring equal representation in every batch. We use SGD optimizer and cosine annealing scheduler during both pretraining and fine-tuning. The learning rate is 0.01 for pretraining and 0.001 for fine-tuning. The default observation ratio $\alpha = 50\%$. Each experiment is repeated 3 times, and we report the average results. For more details on the training process, please refer to Appendix \ref{training_details_}.

\subsection{Comparison Results\label{main_results}}
In this section, we present the main results of our model. We compare TrafficPPT with two state-of-the-art methods: Cam-Traj-Rec \cite{10.1145/3534678.3539186} and Traj2Traj \cite{liao2023traj2traj}. Cam-Traj-Rec stands for prior based methods and Traj2Traj is a deep learning method. We provide the overall Mean Absolute Error (MAE) comparison in \cref{overall_MAE_comparison} and visualize the road volume predictions in \cref{Vol_road,Vol_time}.

\subsubsection{Overall MAE Comparison\label{overall_MAE_comparison}}
\begin{table*}
    \centering
    \caption{Overall MAE comparisons under different checkpoint ratios.\label{tab_MAE_comparison}}
    \resizebox{0.9\linewidth}{!}{
        \begin{tabular}{cccccc|ccccc}
            \toprule
            ~                & \multicolumn{5}{c}{\textbf{Boston}} & \multicolumn{5}{c}{\textbf{Jinan}}                                                                  \\
            \cmidrule(r){2-6} \cmidrule(r){6-11}
            Checkpoint ratio & 10\%                                 & 20\%                                & 30\%  & 50\%   & Time   & 10\%   & 20\%   & 30\%   & 50\%   & Time   \\
            \midrule
            Cam-Traj-Rec     & 7.29                                & 4.08                               & 3.33 & 1.63  & 93.2 s & 0.355 & 0.300 & 0.248 & 0.179 & 83.7 s \\
            Traj2Traj        & 4.71                                & 3.15                               & 2.56 & 1.25  & 41.5 s & 0.297 & 0.249 & 0.232 & 0.143 & 57.6 s \\
            TrafficPPT          & 2.24                                & 1.22                               & 1.07 & 0.667 & 3.35 s & 0.214 & 0.125 & 0.126 & 0.071 & 18.1 s \\
            \bottomrule
        \end{tabular}
        }
\end{table*}
\begin{figure*}
    \centering
    \subfigure[Boston]{\includegraphics[width=0.45\linewidth]{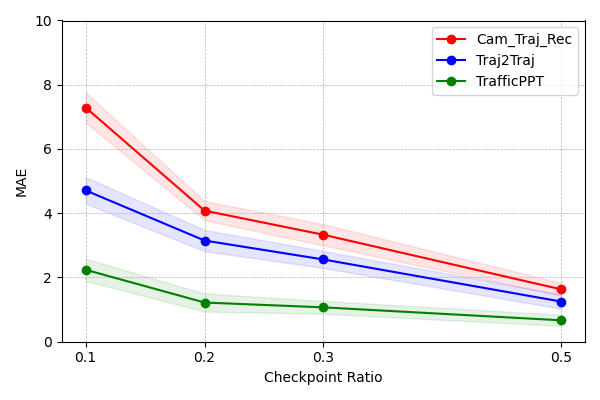}}
    \subfigure[Jinan]{\includegraphics[width=0.45\linewidth]{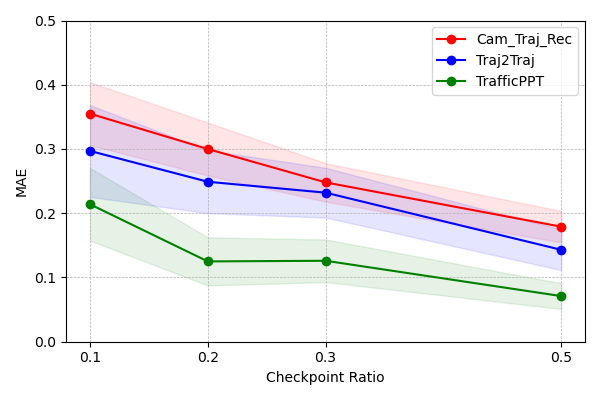}}
    \caption{Overall MAE comparison under different checkpoint ratios.\label{fig_MAE_comparison}}
\end{figure*}
The overall MAE comparison is presented in \cref{tab_MAE_comparison,fig_MAE_comparison}, where we show the MAE of different methods under varying checkpoint ratios. TrafficPPT consistently achieves lower MAE compared to the other two methods. At a checkpoint ratio of 50\%, TrafficPPT's MAE is approximately 50\% lower than that of the other methods. Furthermore, TrafficPPT demonstrates high robustness to low observation ratios. With only a 20\% observation ratio, TrafficPPT already outperforms the other methods at a 50\% observation ratio. In contrast, the end-to-end models Cam-Traj-Rec and Traj2Traj are more sensitive to missing data. Additionally, TrafficPPT exhibits lower time consumption compared to both methods, making it more suitable for real-time applications.

\subsubsection{Volume per Road Comparison\label{Vol_road}}
\begin{figure*}[htbp]
    \centering
    \subfigure[Ground Truth\label{Vol_road_d}]{\includegraphics[width=0.24\linewidth]{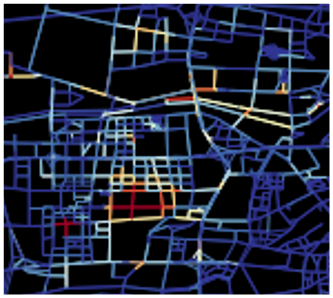}}
    \subfigure[Cam-Traj-Rec\label{Vol_road_a}]{\includegraphics[width=0.24\linewidth]{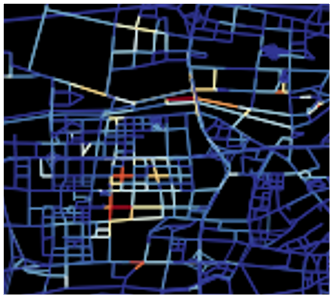}}
    \subfigure[Traj2Traj\label{Vol_road_b}]{\includegraphics[width=0.24\linewidth]{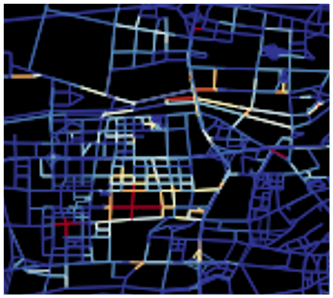}}
    \subfigure[TrafficPPT\label{Vol_road_c}]{\includegraphics[width=0.24\linewidth]{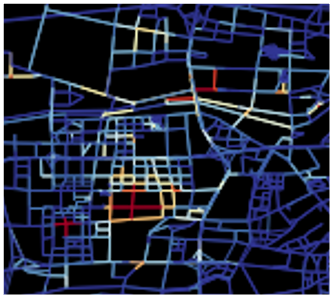}}
    \caption{Volume per road comparison. Blue means low volume and Red means high volume. We focus on the downtown areas, the visualization of the whole city can be found in \cref{full_visualization} and the whole video can be found in supplementary material.\label{fig_Vol_road}}
\end{figure*}

We visualize the traffic volume in the downtown area at the middle time step. As shown in \cref{Vol_road_a}, in the central part of the downtown area, Cam-Traj-Rec exhibits a biased prediction. Cam-Traj-Rec assigns prior distributions to the missing trajectories based on road weights, which are calculated according to the length of different routes between the origin and destination. However, in real world scenarios, vehicles may not always follow the shortest path, leading to misjudgments regarding the busiest roads. In contrast, the deep learning-based method Traj2Traj provides predictions closer to the ground truth, though still lacking in accuracy. As shown in the top-left part of \cref{Vol_road_b}, while Traj2Traj captures the relationships between roads more accurately, the absolute volume predictions remain imprecise. For a more detailed view, please refer to the supplementary material, where we provide videos that show the volume distribution across roads at each time step.

\subsubsection{Volume per Time Step Comparison\label{Vol_time}}

\begin{figure*}[htbp]
    \centering
    \subfigure[Cam-Traj-Rec\label{Vol_time_a}]{\includegraphics[width=0.325\linewidth]{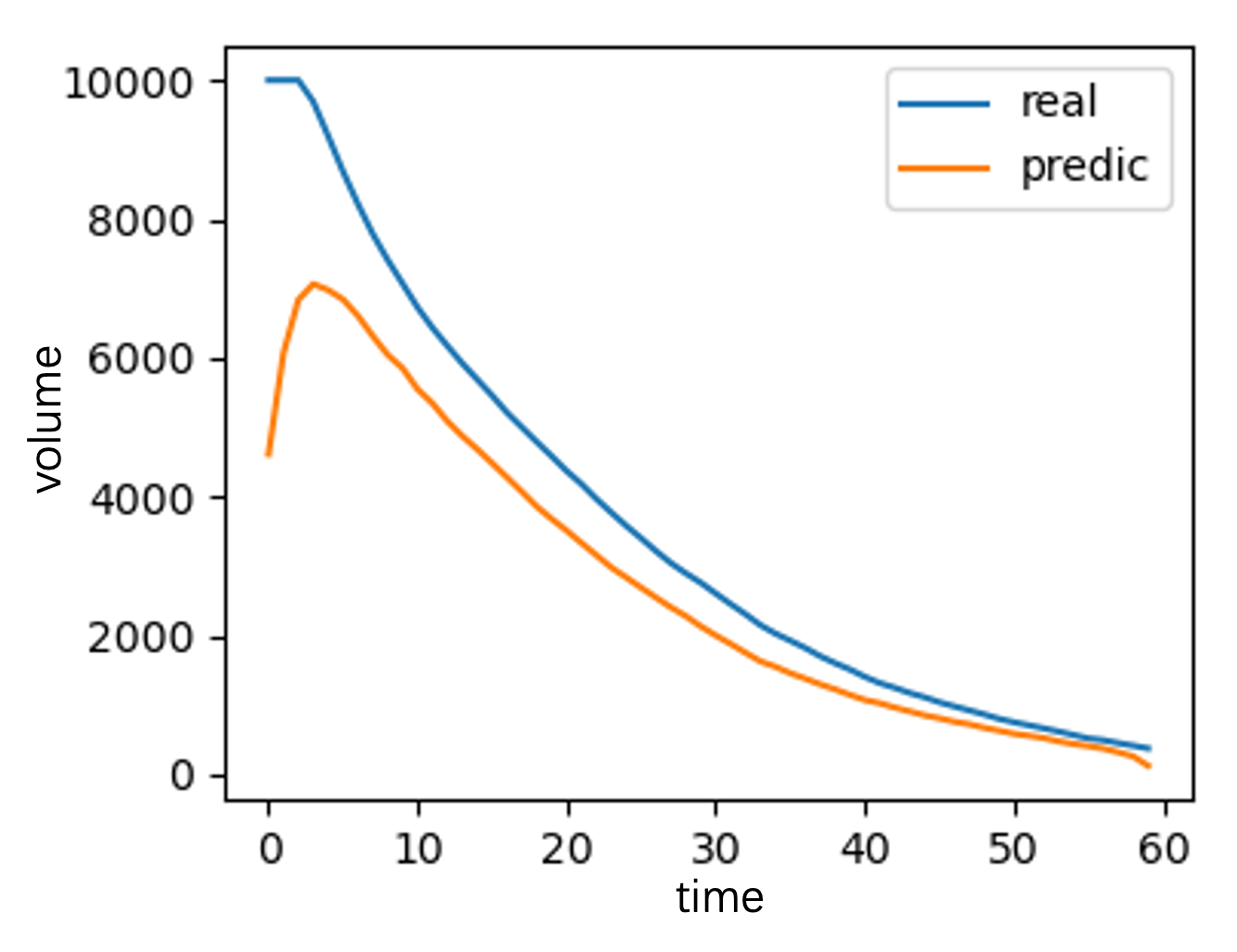}}
    \subfigure[Traj2Traj\label{Vol_time_b}]{\includegraphics[width=0.325\linewidth]{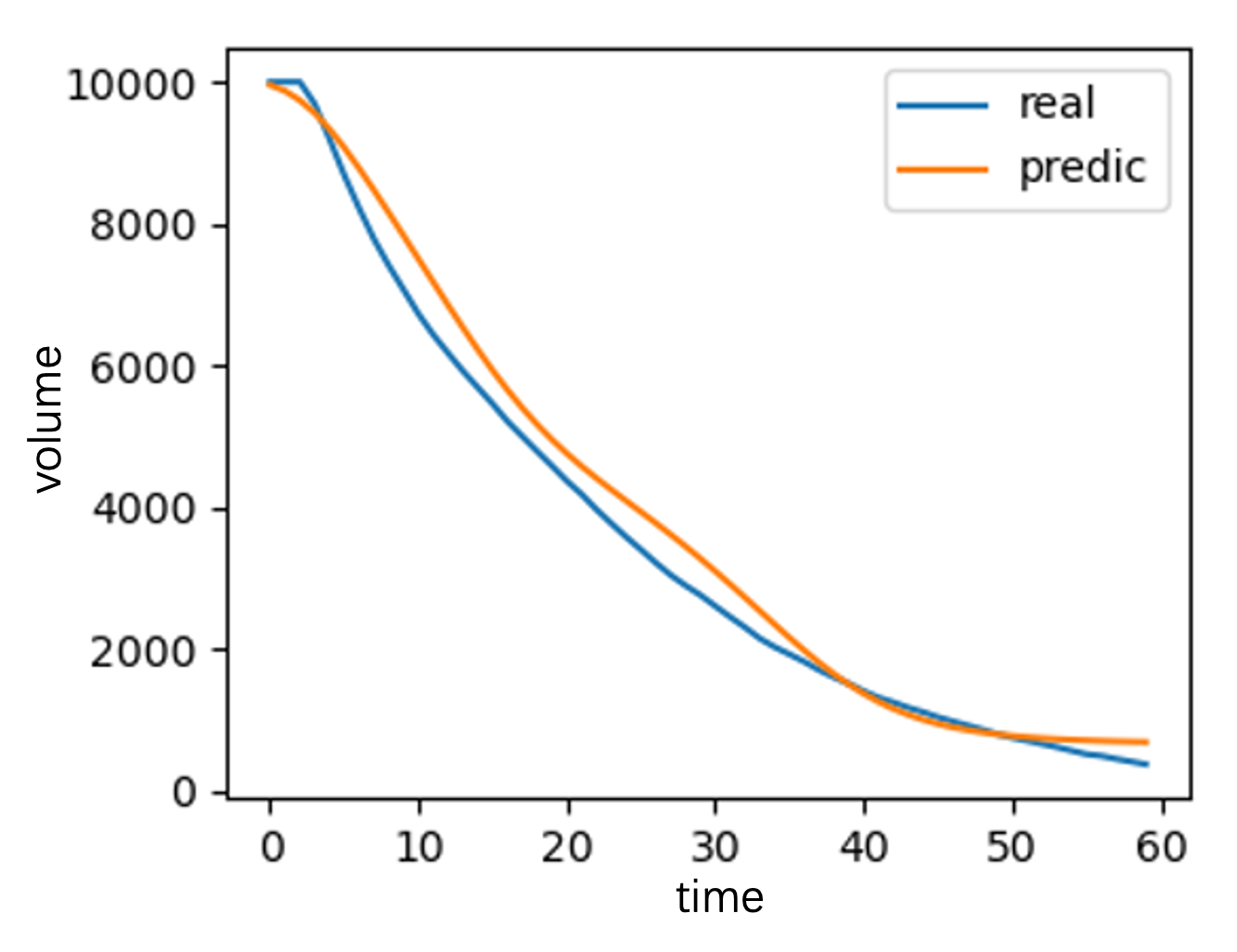}}
    \subfigure[TrafficPPT\label{Vol_time_c}]{\includegraphics[width=0.325\linewidth]{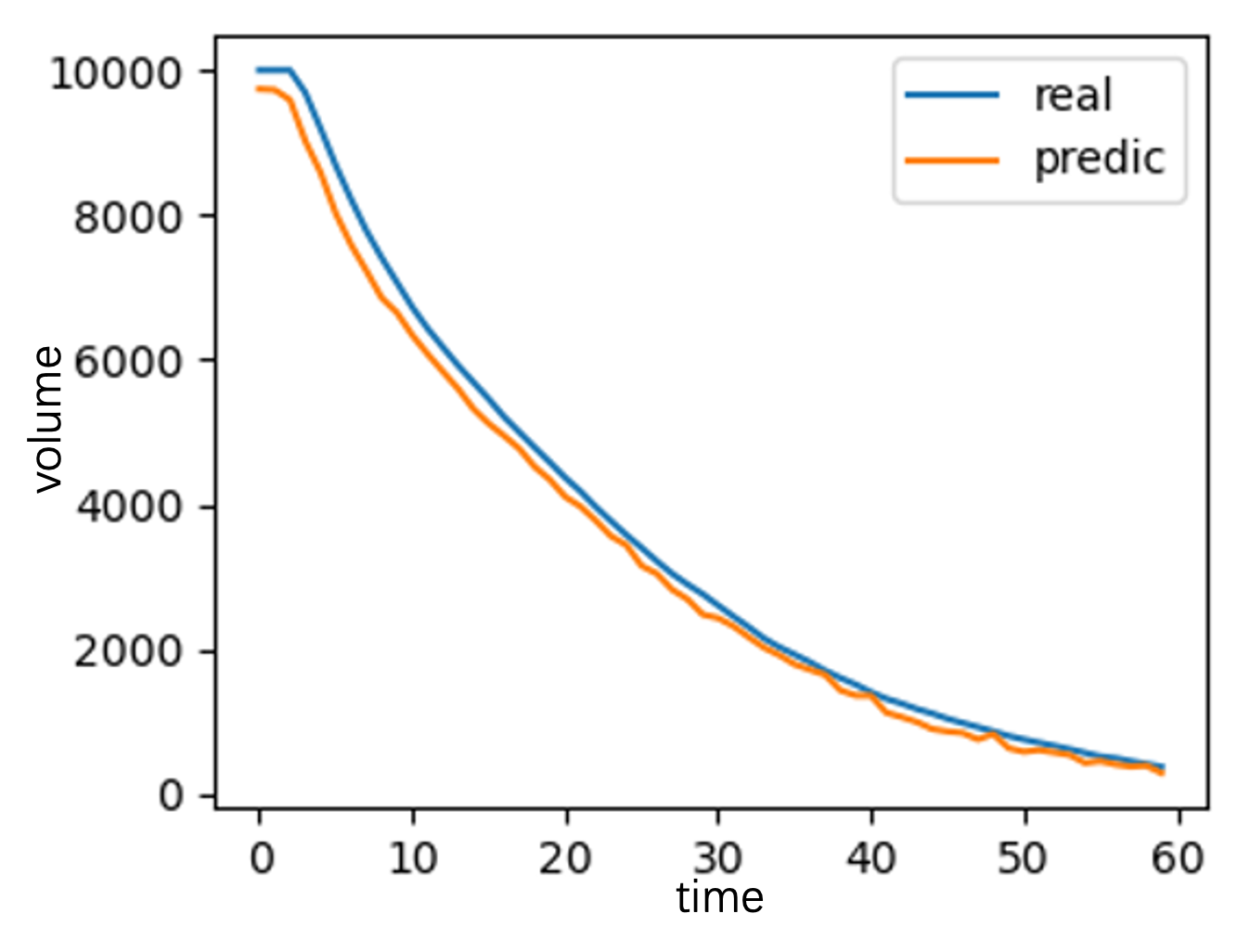}}
    \caption{Volume per time step comparison. The blue curves represent the ground truth volumes across all roads, while the orange curves correspond to the predicted volumes.\label{fig_Vol_time}}
\end{figure*}

We visualize the volume per time step in \cref{fig_Vol_time}. As shown in \cref{Vol_time_a}, Cam-Traj-Rec performs poorly at the beginning. This is a common limitation of prior-based methods, which struggle to handle missing data at the beginning. In contrast, \cref{Vol_time_b} demonstrates that Traj2Traj performs better, but its predictions consistently diverge from the actual curves. This discrepancy can be attributed to the sensitivity of LSTM-based models to long-term dependencies, which becomes problematic in the presence of incomplete observations. Additionally, autoregressive methods face challenges in determining the appropriate endpoint for trajectories, especially when the input data is highly incomplete. In \cref{Vol_time_c}, TrafficPPT exhibits more accurate performance. Its predictions closely align with the ground truth, and errors during the early stages of prediction do not significantly impact subsequent time steps. However, the predictions from TrafficPPT are less stable compared to the other methods, likely due to its non-autoregressive nature.

\subsection{Pretraining and Fine-tuning Analysis\label{pretraining_finetuning_analysis}}
\begin{figure}[htbp]
    \centering
    \includegraphics[width=0.39\linewidth]{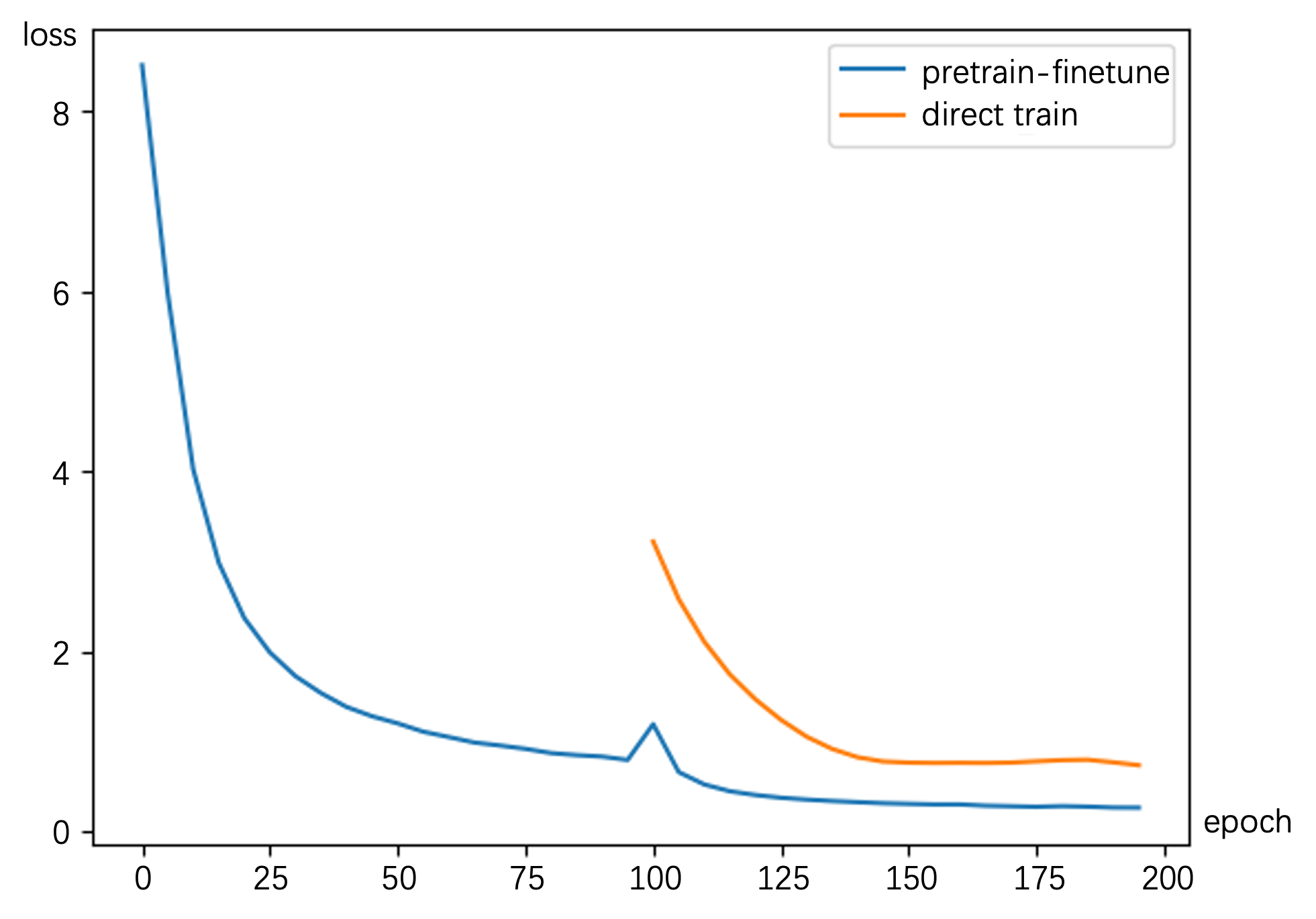}
    \caption{Loss curves of different training mechanism. Blue curve is the pretraining loss and orange curve is the fine-tuning loss. The left part of the blue curve is the pretraining loss and the right part is the fine-tuning loss.\label{fig_pretrain_finetune}}
\end{figure}

To evaluate the effectiveness of the pretraining and fine-tuning mechanism, we compare the loss curves of different training approaches, as shown in \cref{fig_pretrain_finetune}. The left portion of the blue curve represents the loss during the pretraining phase, while the right portion corresponds to the fine-tuning stage. Notably, the fine-tuning curve converges more rapidly and reaches a lower loss than training from scratch, as represented by the orange curve. This result highlights the effectiveness of the pretraining and fine-tuning strategy in enhancing the model's performance. Additionally, a slight increase in the loss is observed in the middle of the orange curve, indicating that direct training may struggle with overfitting, especially when the training data is limited. By allowing the model to learn fundamental patterns and relationships during pretraining, it becomes better equipped to quickly adapt during the fine-tuning phase.

\subsection{Ablation Study\label{ablation_study}}
\begin{table*}[htbp]
    \centering
    \caption{Ablation Study on Boston. ``History'' and ``Adjacency'' means whether we use these modal. ``Discretization'' means whether the continuous inputs are discretized, and to which factor. ``Adj embedding shape'' shows how AvgPooling reshape the adj embeddings. ``Adj attention type'' shows the attention mechanism between the observation embeddings and the adj embeddings.\label{tab_ablation}}
    \resizebox{1\linewidth}{!}{
        \begin{tabular}{cccccccc}
            \toprule
            ~     & \multicolumn{3}{c}{\textbf{Data hyperparameters}} & \multicolumn{2}{c}{\textbf{Structure hyperparameters}} & \multicolumn{2}{c}{\textbf{Performance}}                                                             \\
            \cmidrule(r){2-4} \cmidrule(r){5-6} \cmidrule(r){7-8}
            Index & History                                           & Adjacency                                              & Discretization                           & Adj embedding shape & Adj attention type & MAE   & Time   \\
            \midrule
            (1)   & w/                                                & w/                                                     & w/o                                      & BV1C                & Multi-query        & 0.667 & 3.35 s \\
            \midrule
            (2)   & w/o                                               & w/                                                     & w/o                                      & BV1C                & Multi-query        & 1.12  & 3.01 s \\
            (3)   & w/                                                & w/o                                                    & -                                        & -                   & -                  & 2.21  & 1.13 s \\
            (4)   & w/                                                & w/                                                     & 10                                       & BV1C                & Multi-query        & 0.716 & 2.22 s \\
            (5)   & w/                                                & w/                                                     & 20                                       & BV1C                & Multi-query        & 0.683 & 2.41 s \\
            \midrule
            (6)   & w/                                                & w/                                                     & w/o                                      & BVLC                & Multi-query        & 0.451 & 12.7 s \\
            (7)   & w/                                                & w/                                                     & w/o                                      & B11C                & Multi-query        & 0.894 & 1.77 s \\
            (8)   & w/                                                & w/                                                     & w/o                                      & BV1C                & Multi-head         & 0.538 & 9.50 s \\
            \bottomrule
        \end{tabular}}
\end{table*}

We conduct an ablation study to assess the impact of various We conduct an ablation study to evaluate the impact of various computation-efficient mechanisms. These experiments are limited to the Boston dataset, as the "BVLC" token shape is too large to be applied to the Jinan dataset. The results are summarized in \cref{tab_ablation}. From lines (2) and (3), it is evident that incorporating both historical data and road network information leads to a significant reduction in MAE, with the road network information proving to be more critical. Lines (4) and (5) demonstrate that the discretization mechanism effectively reduces computational overhead without notably compromising performance. Furthermore, lines (6) and (7) highlight that while the token shape of the adjacency table has a negligible effect on model performance, it considerably impacts time complexity. Finally, line (8) illustrates that the multi-query attention mechanism also contributes to reducing computational load while maintaining stable performance.

\section{Discussion}

\subsection{Balance between Performance and Efficiency}
TrafficPPT is designed to achieve a balance between performance and computational efficiency. The multi-view attention mechanism effectively enhance performance; however, it also introduces a substantial computational burden. To mitigate this, mechanisms such as discretization, multi-query attention, and pooling are employed, each slightly compromising performance. While these mechanisms may be optional for smaller road networks, depending on specific requirements, they become essential for the practical deployment of TrafficPPT on larger road networks.

\subsection{Difficulties in Large-scale Pretraining}
TrafficPPT requires predefining the maximum node number across all road networks to ensure that different cities can share the same transformer backbone. However, this approach may lead to numerous padding tokens in smaller cities, potentially affecting performance. Furthermore, the fixed maximum node number limits the model's ability to adapt to larger cities that may not be included in the initial pretraining phase. In future work, we will investigate more flexible architectural designs that can dynamically adjust to different road networks. Additionally, continual learning can be incorporated, enabling the model to learn progressively from collected or simulated data.

\section{Conclusion}

In this paper, we proposed a pretrained probabilistic transformer (TrafficPPT) to address the challenge of city-scale traffic volume prediction with incomplete observations. We introduce a probabilistic framework that predicts traffic volume based on trajectory probabilities. To estimate these probabilities, we design a transformer model that integrates multi-source traffic data. The model is pretrained with simulation data from different cities to enhance the generalizability, followed by fine-tuning with real-world data to tailor it to specific cities. TrafficPPT's one-step road volume prediction, combined with various computation-efficient mechanisms, ensures both high performance and computational efficiency. Experiments on real-world road networks demonstrate that TrafficPPT outperforms other models while requires only 20\% observations. TrafficPPT shows strong adaptability to real-world scenarios with different levels of infrastructure development

\bibliography{Traj_Prob}

\begin{thebibliography}{10}

\bibitem{vlahogianni2014short}
Eleni~I Vlahogianni, Matthew~G Karlaftis, and John~C Golias.
\newblock {{S}hort-{T}erm {T}raffic {F}orecasting: {W}here {W}e {A}re {A}nd {W}here {W}e’re {G}oing}.
\newblock {\em Transportation Research Part C: Emerging Technologies (TRC)}, 2014.

\bibitem{lv2014traffic}
Yisheng Lv, Yanjie Duan, Wenwen Kang, Zhengxi Li, and Fei-Yue Wang.
\newblock {{T}raffic {F}low {P}rediction {W}ith {B}ig {D}ata: {A} {D}eep {L}earning {A}pproach}.
\newblock {\em IEEE Transactions on Intelligent Transportation Systems (T-ITS)}, 2014.

\bibitem{yu2017spatio}
Bing Yu, Haoteng Yin, and Zhanxing Zhu.
\newblock {{S}patio}-{T}emporal {{G}raph} {{C}onvolutional} {{N}etworks}: {{A} {D}eep} {{L}earning} {{F}ramework} {{F}or} {{T}raffic} {{F}orecasting}.
\newblock {\em arXiv}, 2017.

\bibitem{li2017diffusion}
Yaguang Li, Rose Yu, Cyrus Shahabi, and Yan Liu.
\newblock {D}iffusion {C}onvolutional {R}ecurrent {N}eural {N}etwork: {D}ata-{D}riven {T}raffic {F}orecasting.
\newblock {\em arXiv}, 2017.

\bibitem{fang2020meta}
Shen Fang, Xianbing Pan, Shiming Xiang, and Chunhong Pan.
\newblock {{M}eta-{M}{S}{N}et: {M}eta-{L}earning {B}ased {M}ulti-{S}ource {D}ata {F}usion {F}or {T}raffic {F}low {P}rediction}.
\newblock {\em IEEE Signal Processing Letters (SPL)}, 2020.

\bibitem{chen2024traffic}
Jian Chen, Li~Zheng, Yuzhu Hu, Wei Wang, Hongxing Zhang, and Xiping Hu.
\newblock {{T}raffic {F}low {M}atrix-{B}ased {G}raph {N}eural {N}etwork {W}ith {A}ttention {M}echanism {F}or {T}raffic {F}low {P}rediction}.
\newblock {\em Information Fusion (IF)}, 2024.

\bibitem{chen2023vehicle}
Jing Chen, ZhaoChong Zhang, GuoWei Yang, Wei Wang, JiaJia Zhang, and ChunHui Wu.
\newblock {{V}ehicle {F}low {P}rediction {A}t {C}heckpoint {C}onsidering {T}rajectory {B}ased {O}n {C}onvolutional {L}ong {S}hort-{T}erm {M}emory {N}etwork}.
\newblock In {\em Asia Symposium on Image Processing (ASIP)}, 2023.

\bibitem{yu2023city}
Fudan Yu, Huan Yan, Rui Chen, Guozhen Zhang, Yu~Liu, Meng Chen, and Yong Li.
\newblock {{C}ity}-{S}cale {{V}ehicle} {{T}rajectory} {{D}ata} {{F}rom} {{T}raffic} {{C}amera} {{V}ideos}.
\newblock {\em Scientific Data (Sci. Data)}, 2023.

\bibitem{bao2023pket}
Yinxin Bao, Jiali Liu, Qinqin Shen, Yang Cao, Weiping Ding, and Quan Shi.
\newblock {{P}{K}{E}{T}-{G}{C}{N}: {P}rior {K}nowledge {E}nhanced {T}ime-{V}arying {G}raph {C}onvolution {N}etwork {F}or {T}raffic {F}low {P}rediction}.
\newblock {\em Information Sciences (INS)}, 2023.

\bibitem{zhang2019short}
Weibin Zhang, Yinghao Yu, Yong Qi, Feng Shu, and Yinhai Wang.
\newblock {{S}hort-{T}erm {T}raffic {F}low {P}rediction {B}ased {O}n {S}patio-{T}emporal {A}nalysis {A}nd {C}{N}{N} {D}eep {L}earning}.
\newblock {\em Transportmetrica A: Transport Science (Transport. A)}, 2019.

\bibitem{guo2024m}
Xiaoyu Guo, Weiwei Xing, Xiang Wei, Weibin Liu, Jian Zhang, and Wei Lu.
\newblock {{M}-{M}ix: {P}atternwise {M}issing {M}ix {F}or {F}illing {T}he {M}issing {V}alues {I}n {T}raffic {F}low {D}ata}.
\newblock {\em Neural Computing and Applications (NCA)}, 2024.

\bibitem{zhang2017deep}
Junbo Zhang, Yu~Zheng, and Dekang Qi.
\newblock {{D}eep} {{S}patio}-{T}emporal {{R}esidual} {{N}etworks} {{F}or} {{C}itywide} {{C}rowd} {{F}lows} {{P}rediction}.
\newblock In {\em Proceedings of the AAAI Conference on Artificial Intelligence (AAAI)}, 2017.

\bibitem{diao2019dynamic}
Zulong Diao, Xin Wang, Dafang Zhang, Yingru Liu, Kun Xie, and Shaoyao He.
\newblock {{D}ynamic} {{S}patial}-{T}emporal {{G}raph} {{C}onvolutional} {{N}eural} {{N}etworks} {{F}or} {{T}raffic} {{F}orecasting}.
\newblock In {\em Proceedings of the AAAI Conference on Artificial Intelligence (AAAI)}, 2019.

\bibitem{liu2020dynamic}
Lingbo Liu, Jiajie Zhen, Guanbin Li, Geng Zhan, Zhaocheng He, Bowen Du, and Liang Lin.
\newblock {{D}ynamic} {{S}patial}-{T}emporal {{R}epresentation} {{L}earning} {{F}or} {{T}raffic} {{F}low} {{P}rediction}.
\newblock {\em IEEE Transactions on Intelligent Transportation Systems (IEEE T-ITS)}, 2020.

\bibitem{kalander2020spatio}
Marcus Kalander, Min Zhou, Chengzhi Zhang, Hanling Yi, and Lujia Pan.
\newblock {{S}patio}-{T}emporal {{H}ybrid} {{G}raph} {{C}onvolutional} {{N}etwork} {{F}or} {{T}raffic} {{F}orecasting} {{I}n} {{T}elecommunication} {{N}etworks}.
\newblock {\em arXiv}, 2020.

\bibitem{liu2018urban}
Zhidan Liu, Zhenjiang Li, Kaishun Wu, and Mo~Li.
\newblock {{U}rban} {{T}raffic} {{P}rediction} {{F}rom} {{M}obility} {{D}ata} {{U}sing} {{D}eep} {{L}earning}.
\newblock {\em IEEE Network (IEEE Netw.)}, 2018.

\bibitem{patterson2020learning}
Andrew Patterson, Aditya Gahlawat, and Naira Hovakimyan.
\newblock {{L}earning} {{P}robabilistic} {{I}ntersection} {{T}raffic} {{M}odels} {{F}or} {{T}rajectory} {{P}rediction}.
\newblock {\em arXiv}, 2020.

\bibitem{hunter2013path}
Timothy Hunter, Pieter Abbeel, and Alexandre Bayen.
\newblock {T}he {P}ath {I}nference {F}ilter: {M}odel-{B}ased {L}ow-{L}atency {M}ap {M}atching of {P}robe {V}ehicle {D}ata.
\newblock {\em IEEE Transactions on Intelligent Transportation Systems (IEEE T-ITS)}, 2013.

\bibitem{iio2023distribution}
Kentaro Iio, Gulshan Noorsumar, Dominique Lord, and Yunlong Zhang.
\newblock {{O}n} {{T}he} {{D}istribution} {{O}f} {{P}robe} {{T}raffic} {{V}olume} {{E}stimated} {{F}rom} {{T}heir} {{F}ootprints}.
\newblock {\em arXiv}, 2023.

\bibitem{tang2023explainable}
Yuanbo Tang, Zhiyuan Peng, and Yang Li.
\newblock {{E}xplainable {T}rajectory {R}epresentation {T}hrough {D}ictionary {L}earning}.
\newblock In {\em ACM International Conference on Advances in Geographic Information Systems (ACM SIGSPATIAL)}, 2023.

\bibitem{ma2015long}
Xiaolei Ma, Zhimin Tao, Yinhai Wang, Haiyang Yu, and Yunpeng Wang.
\newblock {{L}ong} {{S}hort}-{T}erm {{M}emory} {{N}eural} {{N}etwork} {{F}or} {{T}raffic} {{S}peed} {{P}rediction} {{U}sing} {{R}emote} {{M}icrowave} {{S}ensor} {{D}ata}.
\newblock {\em Transportation Research Part C: Emerging Technologies (Transp. Res. Part C)}, 2015.

\bibitem{zhao2019t}
Ling Zhao, Yujiao Song, Chao Zhang, Yu~Liu, Pu~Wang, Tao Lin, Min Deng, and Haifeng Li.
\newblock {{T}-{GCN}}: {{A} {T}emporal} {{G}raph} {{C}onvolutional} {{N}etwork} {{F}or} {{T}raffic} {{P}rediction}.
\newblock {\em IEEE Transactions on Intelligent Transportation Systems (IEEE T-ITS)}, 2019.

\bibitem{wong2022view}
Conghao Wong, Beihao Xia, Ziming Hong, Qinmu Peng, Wei Yuan, Qiong Cao, Yibo Yang, and Xinge You.
\newblock {{V}iew {V}ertically: {A} {H}ierarchical {N}etwork {F}or {T}rajectory {P}rediction {V}ia {F}ourier {S}pectrums}.
\newblock In {\em European Conference on Computer Vision (ECCV)}, 2022.

\bibitem{huang2023reconstructing}
Yuzhu Huang, Awad Abdelhalim, Anson Stewart, Jinhua Zhao, and Haris Koutsopoulos.
\newblock {{R}econstructing {T}ransit {V}ehicle {T}rajectory {U}sing {H}igh-{R}esolution {G}{P}{S} {D}ata}.
\newblock In {\em IEEE International Conference on Intelligent Transportation Systems (ITSC)}, 2023.

\bibitem{achiam2023gpt}
Josh Achiam, Steven Adler, Sandhini Agarwal, Lama Ahmad, Ilge Akkaya, Florencia~Leoni Aleman, Diogo Almeida, Janko Altenschmidt, Sam Altman, Shyamal Anadkat, et~al.
\newblock {{G}pt-4 {T}echnical {R}eport}.
\newblock {\em arXiv preprint arXiv:2303.08774}, 2023.

\bibitem{liu2024deepseek}
Aixin Liu, Bei Feng, Bing Xue, Bingxuan Wang, Bochao Wu, Chengda Lu, Chenggang Zhao, Chengqi Deng, Chenyu Zhang, Chong Ruan, et~al.
\newblock {{D}eepseek-v3 {T}echnical {R}eport}.
\newblock {\em arXiv preprint arXiv:2412.19437}, 2024.

\bibitem{devlin2019bert}
Jacob Devlin, Ming-Wei Chang, Kenton Lee, and Kristina Toutanova.
\newblock {{B}ert: {P}re-Training {O}f {D}eep {B}idirectional {T}ransformers {F}or {L}anguage {U}nderstanding}.
\newblock In {\em Proceedings of the 2019 Conference of the North American Chapter of the Association for Computational Linguistics: Human Language Technologies, Volume 1 (Long and Short Papers)}, 2019.

\bibitem{radford2018improving}
Alec Radford, Karthik Narasimhan, Tim Salimans, Ilya Sutskever, et~al.
\newblock {{I}mproving {L}anguage {U}nderstanding {B}y {G}enerative {P}re-Training}.
\newblock 2018.

\bibitem{he2022masked}
Kaiming He, Xinlei Chen, Saining Xie, Yanghao Li, Piotr Doll{\'a}r, and Ross Girshick.
\newblock {{M}asked {A}utoencoders {A}re {S}calable {V}ision {L}earners}.
\newblock In {\em Proceedings of the IEEE/CVF Conference on Computer Vision and Pattern Recognition}, 2022.

\bibitem{hu2019strategies}
Weihua Hu, Bowen Liu, Joseph Gomes, Marinka Zitnik, Percy Liang, Vijay Pande, and Jure Leskovec.
\newblock {{S}trategies {F}or {P}re-Training {G}raph {N}eural {N}etworks}.
\newblock {\em arXiv preprint arXiv:1905.12265}, 2019.

\bibitem{hu2020gpt}
Ziniu Hu, Yuxiao Dong, Kuansan Wang, Kai-Wei Chang, and Yizhou Sun.
\newblock {{G}pt-Gnn: {G}enerative {P}re-Training {O}f {G}raph {N}eural {N}etworks}.
\newblock In {\em Proceedings of the 26th ACM SIGKDD International Conference on Knowledge Discovery \& Data Mining}, 2020.

\bibitem{hou2024masked}
Lu~Hou, Yunxin Geng, Lingyi Han, Haojun Yang, Kan Zheng, and Xianbin Wang.
\newblock {{M}asked {T}oken {E}nabled {P}re-Training: {A} {T}ask-Agnostic {A}pproach {F}or {U}nderstanding {C}omplex {T}raffic {F}low}.
\newblock {\em IEEE Transactions on Mobile Computing}, 2024.

\bibitem{zhou2024trafficformer}
Guangmeng Zhou, Xiongwen Guo, Zhuotao Liu, Tong Li, Qi~Li, and Ke~Xu.
\newblock {{T}rafficformer: {A}n {E}fficient {P}re-Trained {M}odel {F}or {T}raffic {D}ata}.
\newblock In {\em 2025 IEEE Symposium on Security and Privacy (SP)}, 2024.

\bibitem{jin2021trafficbert}
KyoHoon Jin, JeongA Wi, EunJu Lee, ShinJin Kang, SooKyun Kim, and YoungBin Kim.
\newblock {{T}rafficBERT: {P}re-Trained {M}odel {W}ith {L}arge-Scale {D}ata {F}or {L}ong-Range {T}raffic {F}low {F}orecasting}.
\newblock {\em Expert Systems with Applications}, 2021.

\bibitem{ainslie2023gqa}
Joshua Ainslie, James Lee-Thorp, Michiel de~Jong, Yury Zemlyanskiy, Federico Lebr{\'o}n, and Sumit Sanghai.
\newblock {{GQA}}: {{T}raining} {{G}eneralized} {{M}ulti}-{Q}uery {{T}ransformer} {{M}odels} {{F}rom} {{M}ulti}-{H}ead {{C}heckpoints}.
\newblock {\em arXiv}, 2023.

\bibitem{openstreetmap}
{OpenStreetMap contributors}.
\newblock {OpenStreetMap}.
\newblock \url{https://www.openstreetmap.org/copyright}, 2023.

\bibitem{boeing2017osmnx}
Geoff Boeing.
\newblock {{O}smnx: {N}ew {M}ethods {F}or {A}cquiring, {C}onstructing, {A}nalyzing, {A}nd {V}isualizing {C}omplex {S}treet {N}etworks}.
\newblock {\em Computers, Environment and Urban Systems}, 2017.

\bibitem{liang2023cblab}
Chumeng Liang, Zherui Huang, Yicheng Liu, Zhanyu Liu, Guanjie Zheng, Hanyuan Shi, Kan Wu, Yuhao Du, Fuliang Li, and Zhenhui~Jessie Li.
\newblock {{C}blab: {S}upporting {T}he {T}raining {O}f {L}arge-Scale {T}raffic {C}ontrol {P}olicies {W}ith {S}calable {T}raffic {S}imulation}.
\newblock In {\em Proceedings of the 29th ACM SIGKDD Conference on Knowledge Discovery and Data Mining}, 2023.

\bibitem{10.1145/3534678.3539186}
Fudan Yu, Wenxuan Ao, Huan Yan, Guozhen Zhang, Wei Wu, and Yong Li.
\newblock {{S}patio}-{T}emporal {{V}ehicle} {{T}rajectory} {{R}ecovery} {{O}n} {{R}oad} {{N}etwork} {{B}ased} {{O}n} {{T}raffic} {{C}amera} {{V}ideo} {{D}ata}.
\newblock In {\em Proceedings of the ACM SIGKDD International Conference on Knowledge Discovery and Data Mining (KDD)}, 2022.

\bibitem{liao2023traj2traj}
Lyuchao Liao, Yuyuan Lin, Weifeng Li, Fumin Zou, and Linsen Luo.
\newblock {{T}raj2{T}raj}: {{A} {R}oad} {{N}etwork} {{C}onstrained} {{S}patiotemporal} {{I}nterpolation} {{M}odel} {{F}or} {{T}raffic} {{T}rajectory} {{R}estoration}.
\newblock {\em Transactions in GIS (Trans. GIS)}, 2023.

\end{thebibliography}
\bibliographystyle{unsrt}

\appendix
\newpage

\section{Details of Model Architecture\label{model_structure_}}

\subsection{Overall Architecture\label{overall_arch_}}
\begin{figure*}[htb]
    \centering
    \includegraphics[width=0.9\linewidth]{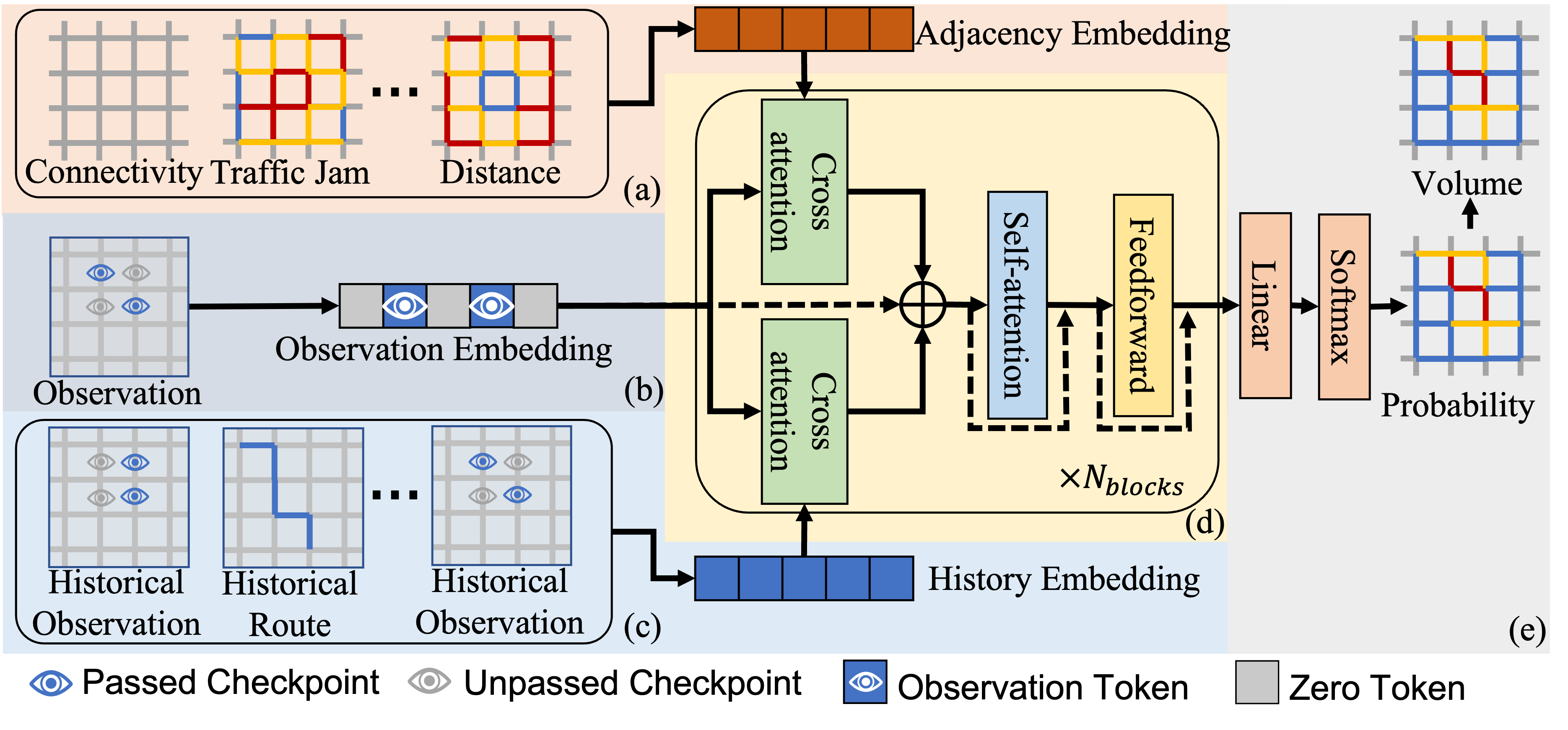}
    \caption{Overview of the proposed TrafficPPT. (a, b, c) represent the embedding layers for road network data, observation data, and historical data, respectively. (d) denotes the multi-view attention block. (e) corresponds to the output projection.\label{fig_overall_arch_}}
\end{figure*}

The overall architecture of our model is shown in \cref{fig_overall_arch_}. The model consists of three components: embedding layers, multi-view attention blocks, and linear projections. The embedding layers project the observed trajectories, road network, and historical trajectories into the same hidden space. These different sources of embeddings are then aggregated within the multi-view attention block, which captures complex relationships across different views. The output from the multi-view attention block is passed through a linear projection layer with softmax activation to compute the trajectory probabilities. Finally, the trajectory probabilities are used to estimate the road volume.

\subsection{Embedding Layers\label{embedding_layers_}}
\begin{figure*}[htb]
    \centering
    \includegraphics[width=0.8\linewidth]{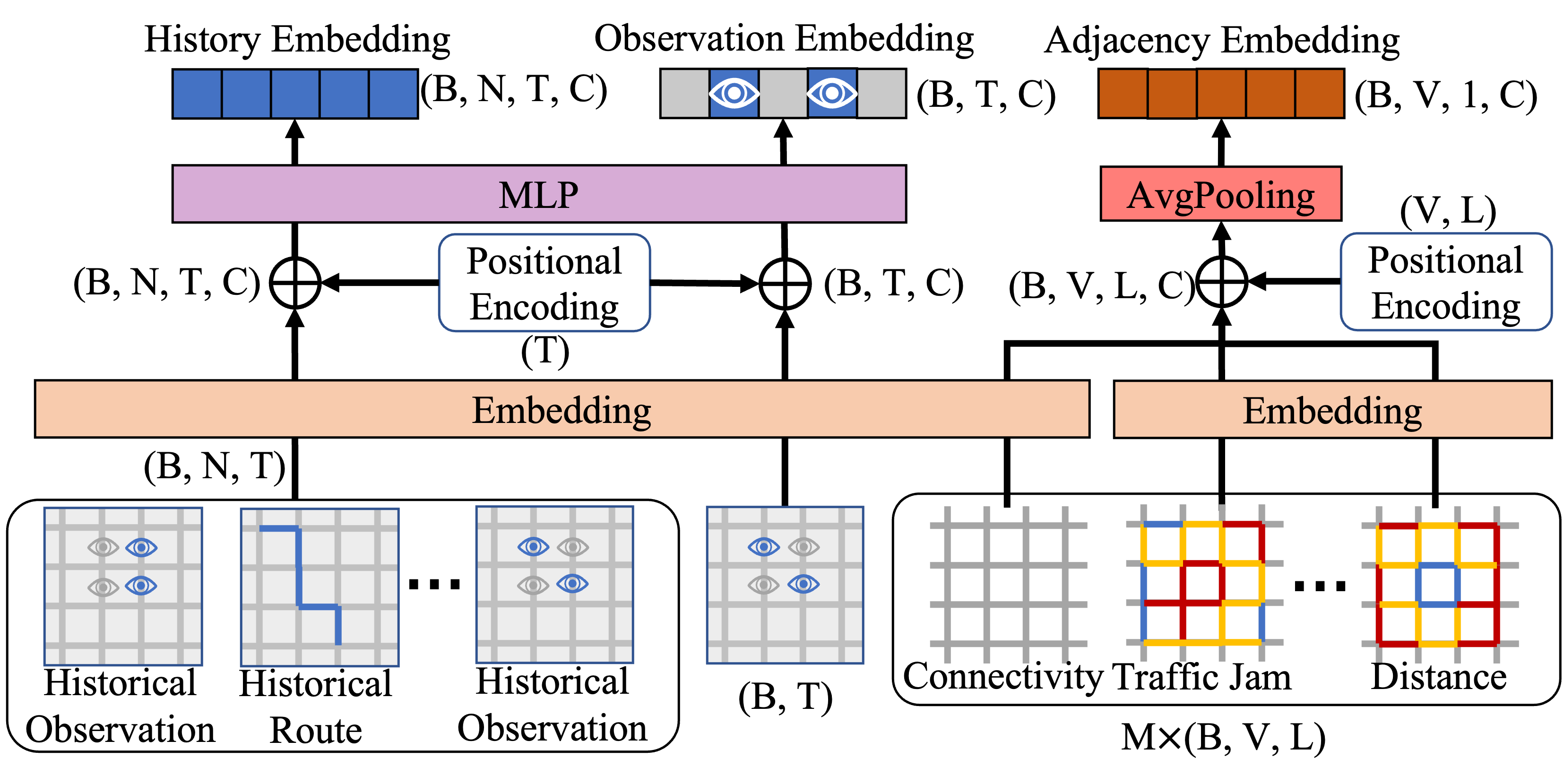}
    \caption{Embedding layers of TrafficPPT.\label{fig_embedding_layers_}}
\end{figure*}

As shown in \cref{fig_embedding_layers_}, the embedding layers project all the real-world information into the aligned token space. The embedding layers consist of three parts: observation embedding, history embedding and road network embedding. The observation embedding and history embedding share the same weights, while the road network needs other embedding tables.

Given a batch of observed trajectories $X \in \{0,1,...,V\}^{B,T}$, we first project the discrete node into continuous tokens by an embedding matrix $E_{traj} \in \mathbb{R}^{V+1, C}$, where $C$ is the dimension of the latent space. Combined with the positional encoding over the time steps, the tokens are further passed through an MLP layer to get the final observation tokens. The MLP layer consists of a linear projection, a layer normalization, and a SiLU activation function. The output of the MLP layer is denoted as $z_{obs} \in \mathbb{R}^{B,T,C}$. To be specific, the observation embedding is calculated as follows:
\begin{align}
    z_{obs} = \text{SiLU}(\text{LN}(\text{Linear}(E_{traj}(X) + E_{pos}^T)))
\end{align}
where LN denotes LayerNorm. $E_{pos}^T \in \mathbb{R}^{B,T,C}$ is the positional encoding over T.

The historical data consist of past trajectories, which may be either complete or incomplete. Incomplete trajectories refer to past observations of the target vehicle recorded at checkpoints. Complete trajectories, on the other hand, can be obtained from sources such as GPS data, past trajectory probability estimations, or other relevant datasets. If no historical data are available, they are set to zeros. Since historical data are also in the form of trajectories, they are processed similarly to the observation data. The history embedding is computed as follows:
\begin{align}
    z_{his} = \text{SiLU}(\text{LN}(\text{Linear}(E_{traj}(X_{his}) + N\times E_{pos}^T)))
\end{align}
where $X_{his} \in \{0,1,...,V\}^{B,N,T}$ is the historical data, $N$ is the number of historical trajectories. $E_{pos}^T$ is repeat $N$ times to match the dimension.

The road network data are represented as adjacency tables, denoted as $A=[A_0, A_1,...,A_M]$, where $M$ is the number of adjacency tables. $A_0 \in \{0,1,...,V\}^{B,V,L}$ is a special adjacency table that represents the road network connections, where $L$ is the max connectivity. $A_0[b,v,l]$ is the $l_{th}$ neighbor of node $v$ for the $b_{th}$ trajectory, and 0 indicates no additional neighbors. The other adjacency tables provide information about roads, such as speed limits, distances, or road types, and are aligned with $A_0$ to ensure that $A_i[b, v, l]$ corresponds to the same road as $A_0[b, v, l]$. If $A_i$ contains continuous data, it can be projected into tokens by MLP layers; if $A_i$ contains discrete data, an embedding matrix is used. To reduce computation, the continuous data could be discretized into several bins, and an embedding matrix is then applied. The road network embedding is calculated as follows:
\begin{align}
    z_{adj} = \text{AvgPooling}(E_{traj}(A_0) + \sum_{i = 1}^{M}E_{adj}(A_i) + E_{pos}^{V, L})
\end{align}
where $A_i \in \{1,...,K\}^{V,L}$ is the discretized road weights, $K$ is the discretization level, $E_{adj} \in \mathbb{R}^{K, C}$ is the embedding matrix of the adjacency tables, $E_{pos}^{V, L} \in \mathbb{R}^{B, V, L, C}$ is the positional encoding over (V, L). We use average pooling to reduce the computation load.

\subsection{Multi-view Attention Block\label{multi_view_attention_block_}}
\begin{figure*}[htb]
    \centering
    \includegraphics[width=0.8\linewidth]{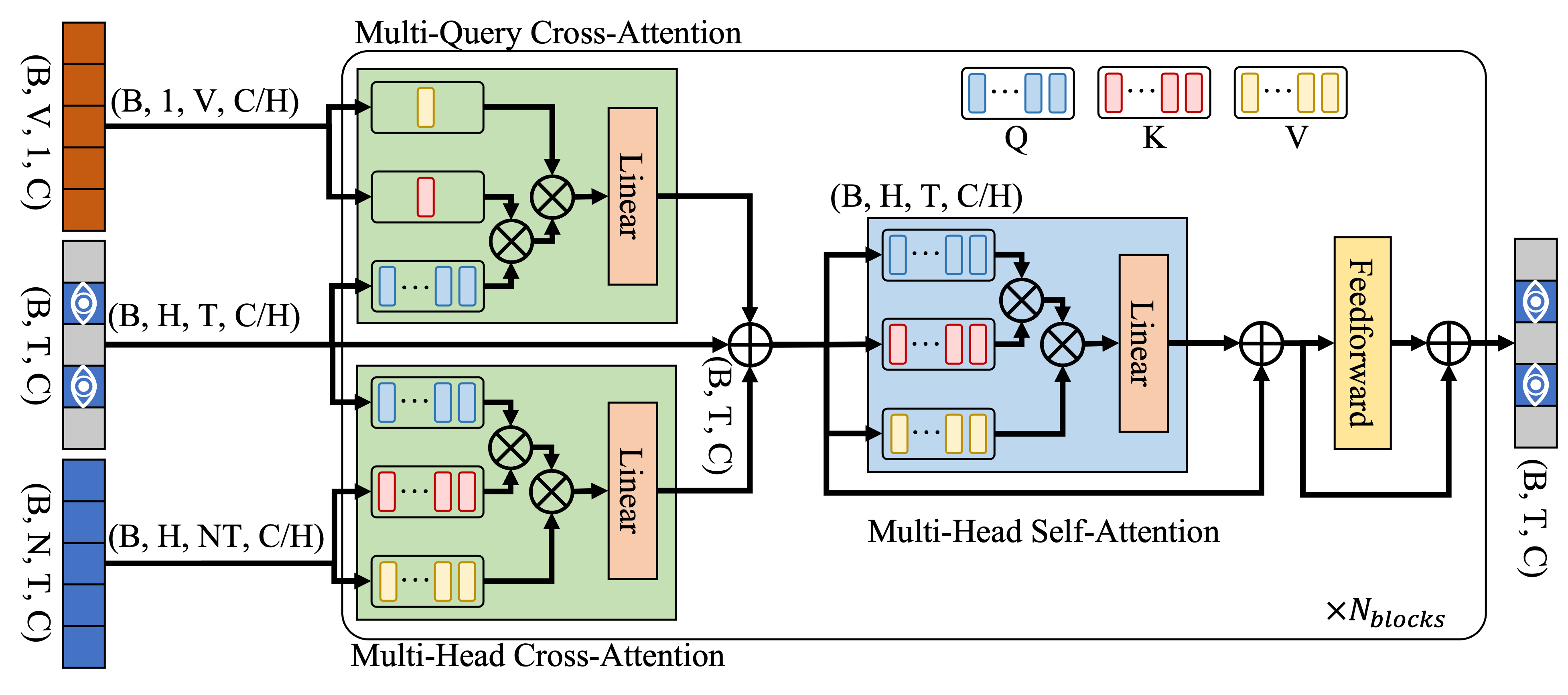}
    \caption{Multi-view attention block of TrafficPPT.\label{fig_multi_view_attention_block_}}
\end{figure*}

The multi-view attention block is the core component of our model, which integrates the observation information, history information, and road network information. As shown in \cref{fig_multi_view_attention_block_}, the multi-view attention block consists of two cross-attention blocks and one self-attention block, each wrapped with residual connections and LayerNorm. We adopt multi-head attention and multi-query attention \cite{ainslie2023gqa} mechanism to capture the complex relationship between different views. The adjacency tokens represent the information of each node, which is comprehensive but not efficient. To address this problem, we adopt the multi-query attention mechanism to reduce the computation. The adjacency tokens are linearly projected into the key and value tokens, while the observation tokens are linearly projected into the query tokens. As demonstrated in our ablation study, the multi-query attention mechanism effectively reduces computation with minimal impact on model performance. The multi-view attention block is computed as follows:
\begin{align}
    \begin{aligned}
        z_{adj}^{'}  & = \text{MQA}(z_{obs}, z_{adj}, z_{adj})                           \\
        z_{his}^{'}  & = \text{MHA}(z_{obs}, z_{his}, z_{his})                           \\
        z_{obs}^{'}  & = z_{obs} + z_{adj}^{'} + z_{his}^{'}                             \\
        z_{obs}^{'}  & = \text{MQA}(z_{obs}^{'}, z_{obs}^{'}, z_{obs}^{'}) + z_{obs}^{'} \\
        z_{obs}^{''} & = \text{FFN}(z_{obs}^{'})+ z_{obs}^{'}
    \end{aligned}
\end{align}
where MQA and MHA denote the multi-query attention and multi-head attention, which project the inputs into query, key, and value tokens accordingly. FFN denotes the feed-forward network, which consists of two linear projections and a SiLU activation function. $z_{adj}^{'},z_{his}^{'}, z_{obs}^{'}$ are intermediate tokens and $z_{obs}^{''}$ is the output of the multi-view attention block.

After the multi-view attention block, the hidden tokens are projected into the trajectory probability by a linear projection and a softmax activation function. The output is calculated as follows:
\begin{align}
    q_\theta(Y|X,X_{hist},A) = \text{Softmax}(\text{Linear}(z_{obs}^{''}))
\end{align}
where $\theta$ is the parameter of our model, $Y \in [0,1]^{B,T,V}$ is the estimated trajectory probability.

\section{Training Details\label{training_details_}}
\begin{table*}[h!]
    \centering
    \caption{Default hyperparameters.\label{tab_hyperparameters_}}
    \resizebox{1\linewidth}{!}{
        \begin{tabular}{ccccccccccccc}
            \toprule
            ~       & \multicolumn{4}{c}{\textbf{Model Parameters}} & \multicolumn{4}{c}{\textbf{Pretraining Parameters}}& \multicolumn{4}{c}{\textbf{Fine-tuning Parameters}}                                   \\
            \cmidrule(r){2-5} \cmidrule(r){6-9} \cmidrule{10-13}
            Dataset & Blocks                                        & \makecell{Hidden                                                                   \\size} & Heads & \makecell{FFN expansion\\factor} & \makecell{Discretization\\factor}  & \makecell{Batch\\size} & Lr & Epochs& \makecell{Discretization\\factor}  & \makecell{Batch\\size} & Lr & Epochs\\
            \midrule
            Boston  & \multirow{2}{*}{8}                                             & \multirow{2}{*}{64}                                               & \multirow{2}{*}{16} & \multirow{2}{*}{2} & w/o & \multirow{2}{*}{50} & \multirow{2}{*}{0.01} & \multirow{2}{*}{100}& w/o & 512 & 0.001 & 100  \\
            Jinan   & ~                                             & ~                                               & ~ & ~ & 30  & ~  & ~ & ~ & 30  & 50  & 0.001 & 100\\
            \bottomrule
        \end{tabular}}
\end{table*}
The experiments are conducted on a server equipped with 4 NVIDIA A30 GPUs. We use Stochastic Gradient Descent (SGD) as the optimizer, along with a Cosine Annealing learning rate scheduler. The default hyperparameters are presented in \cref{tab_hyperparameters_}. For the Boston road network, being relatively small, we utilize an MLP as the tokenizer rather than employing the discretization mechanism. For the larger Jinan road network, we apply the discretization mechanism to reduce computational overhead. Since most trajectories do not reach the maximum of 60 steps, we apply a mask over the loss function to ignore the padding steps. The masked value for the loss between the prediction and padding steps is set to 0.0001, ensuring that the training primarily focuses on real steps while allowing the output trajectories to terminate at the padding steps. The pretraining takes approximately 200 GPU hours. The training for the Boston dataset takes approximately 8 GPU hours, while training on the Jinan dataset takes about 100 GPU hours. Each training process is repeated 3 times, and we report the average results. During each training iteration, the checkpoints are randomly selected with the default ratio $\alpha = 0.5$.


\newpage
\section{Full Visualization of the Volume Distribution\label{full_visualization}}
\begin{figure*}[h!]
    \centering
    \subfigure[Ground Truth]{\includegraphics[width=0.8\linewidth]{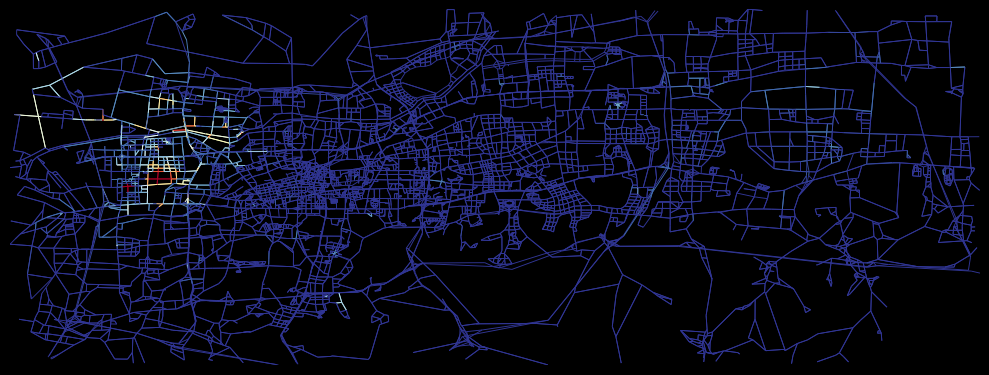}}
    \subfigure[Cam-Traj-Rec]{\includegraphics[width=0.8\linewidth]{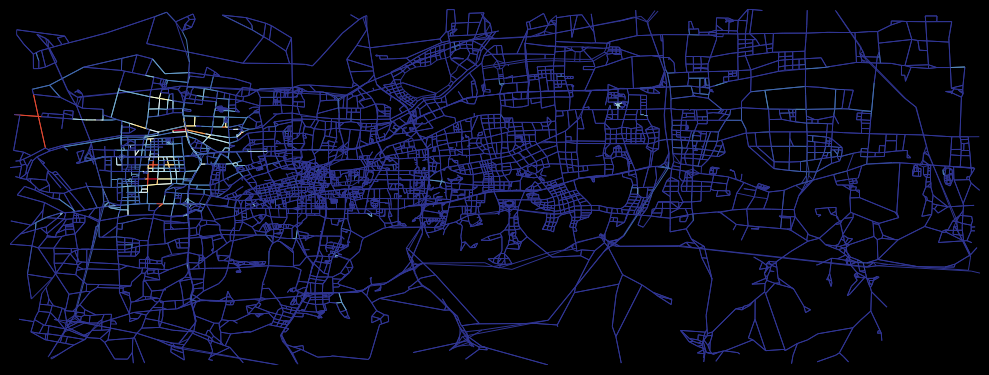}}
    \subfigure[Traj2Traj]{\includegraphics[width=0.8\linewidth]{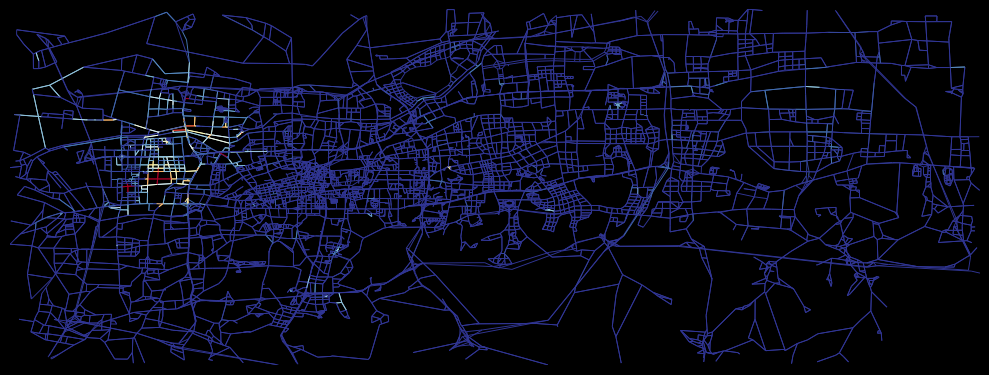}}
    \subfigure[TrafficPPT]{\includegraphics[width=0.8\linewidth]{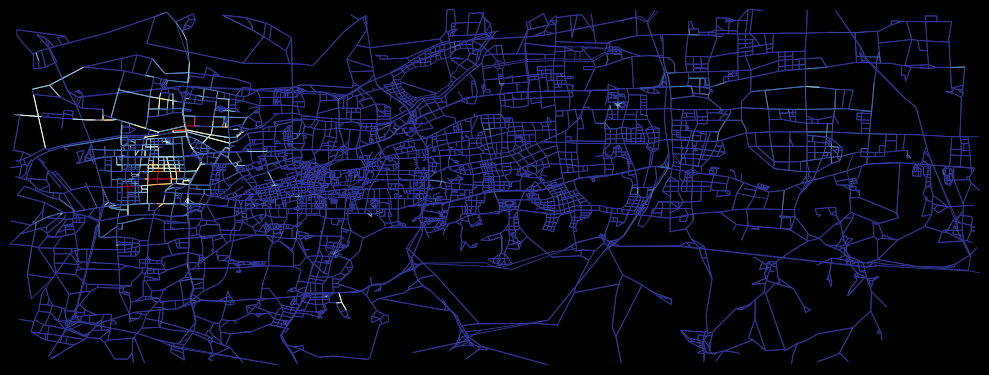}}
    \caption{Volume per road comparison. Blue means low volume and Red means high volume.}
\end{figure*}

\end{document}